\documentclass[10pt,twocolumn,letterpaper]{article}
\usepackage[pagenumbers]{cvpr}
\usepackage{microtype}
\usepackage[T1]{fontenc}
\usepackage{tabularx}

\newcommand{\method}{WA-JEPA}
\newcommand{\sg}{\operatorname{sg}}
\newcommand{\mse}{\operatorname{MSE}}
\newcommand{\slone}{\operatorname{SmoothL1}}

\usepackage{placeins}
\definecolor{cvprblue}{rgb}{0.21,0.49,0.74}
\usepackage[breaklinks,colorlinks,allcolors=cvprblue]{hyperref}
\hypersetup{pdftitle={WA-JEPA: Goal-Aware Predictive Feedback Action Refinement for a World-Action Model},pdfauthor={Tao An, Weihang Liang, Tianqi Wang, Chuqing Cao, Wenming Tan, Kai Zhang, Haibo Lei, Tao Xie, Lijun Zhao, Ruifeng Li}}
\title{WA-JEPA: Goal-Aware Predictive Feedback Action Refinement\\for a World--Action Model}
\author{%
Tao An$^{1}$, Weihang Liang$^{1}$, Tianqi Wang$^{1}$, Chuqing Cao$^{2}$, Wenming Tan$^{3}$,\\
Kai Zhang$^{3}$, Haibo Lei$^{3}$, Tao Xie$^{1}$, Lijun Zhao$^{1}$, Ruifeng Li$^{1}$\thanks{Corresponding author.}\\[4pt]
{\normalsize $^{1}$Harbin Institute of Technology}\\
{\normalsize $^{2}$Anhui Polytechnic University}\qquad
{\normalsize $^{3}$Hikvision Research Institute}
}

\begin{document}
\maketitle
\raggedbottom
\begin{abstract}
JEPA-based world models predict action consequences in latent space, but representative planning approaches rely on candidate-action search, requiring repeated evaluation of action sequences and incurring substantial online computational cost. Direct action generation avoids this search, yet how to leverage latent predictions to guide action generation given task goals remains a key challenge in translating learned physical dynamics into effective decisions.
We introduce \method{}, a world--action model that couples action generation, action-conditioned latent prediction, and goal-aware feedback. The model first generates an action proposal and predicts the resulting future scenes and proprioceptive states, then refines the proposal through two complementary feedback paths, using the task goal as a reference.
Goal-discrepancy residual feedback (GDRF) uses differences between the latent representations of predicted scenes and the task goal to guide chunk-wise action refinement. Future-scene context feedback (FSCF) preserves fine-grained predicted scene representations to guide subsequent action refinement. Jointly learned from offline demonstrations, this process iteratively updates the same plan before execution, making future predictions directly participate in action generation without candidate-action search. We evaluate short-horizon goal reaching, long-horizon tasks, and full-task execution. The results show that goal-aware predictive feedback supports competitive task performance while maintaining a compact model and low online computational cost.
Code is available at \url{https://github.com/Dovisya/WA-JEPA}.
\end{abstract}

\section{Introduction}
\label{sec:intro}
Robot control requires generating actions from observations and task goals to bring the environment toward a desired state. World models learn action-conditioned dynamics to predict how actions change the environment, informing decisions before execution~\cite{zhou2024dinowm,maes2026lewm}. Research based on joint-embedding predictive architectures (JEPAs) models these changes in latent space without reconstructing future images~\cite{assran2025vjepa2}. DINO-WM learns action-conditioned dynamics over pretrained visual features, whereas LeWM jointly learns a visual encoder and latent predictor~\cite{zhou2024dinowm,maes2026lewm}. These models allow predicted scenes and task goals to be compared in a common feature space, providing a basis for visual goal-directed control.

Translating latent predictions into executable actions is central to control with these world models. Methods such as LeWM use candidate-action search: a planner repeatedly samples action sequences, predicts their consequences, and updates the sampling distribution according to the distance between predicted outcomes and the goal. This approach uses learned dynamics to select actions, but repeated candidate evaluation incurs substantial online computation. Imitation learning instead learns action generation from offline demonstrations and directly outputs control sequences without candidate search~\cite{zhao2023act,chi2023diffusion,zhou2026patchpolicy}. However, these policies do not explicitly use predicted action consequences when generating control sequences. World–action models jointly model future observations and actions, allowing predictive learning to support action generation~\cite{wu2024gr1,ye2026dreamzero}. Building on these advances, we investigate how predictions of action consequences can improve goal attainment while retaining the computational efficiency of direct action generation. We explore goal-aware predictive feedback as an approach to this problem: the model compares the predicted outcomes of its current action proposal with the task goal in JEPA latent space and uses this information to iteratively refine the same plan before execution, without candidate-action search.

We introduce \method{}, a world--action model that couples action generation, action-conditioned latent prediction, and goal-aware feedback. Given observation history, proprioception, and the task goal, the model first generates an action proposal, and the world model predicts the resulting future scenes and proprioceptive states. \emph{Goal-discrepancy residual feedback (GDRF)} compares the predicted scene for each chunk with the task goal. It combines this discrepancy with the current action representation and predicted state to learn action-token corrections, translating goal discrepancies in visual representations into guidance for action refinement. \emph{Future-scene context feedback (FSCF)} preserves predicted scene representations, providing context for subsequent action refinement beyond goal discrepancies. Together, the two paths combine goal-directed chunk-wise corrections with detailed future-scene context. Action and prediction supervision from offline demonstrations jointly train the model to cycle through action generation, future prediction, and feedback correction, repeatedly updating the same pending plan before execution. The process uses a fixed number of forward stages, making predictions part of deployment-time action optimization without candidate-action search. It thus grounds goal-directed action refinement in learned relationships between actions and changes in the physical environment.

We evaluate \method{} at three levels: short-horizon goal reaching, long-horizon tasks, and full-task execution. Following LeWM's short-horizon protocol, we compare against world-model planners, imitation-learning methods, and offline reinforcement-learning methods. \method{} achieves strong goal-reaching performance, matching or improving LeWM's success rates while accelerating complete-plan generation by $10.8$--$32.6\times$. For long-horizon tasks, we integrate \method{} into the Anchored Planning framework of \emph{Aim Short to Reach Far}~\cite{liu2026aimshort} without retraining. High-level retrieval supplies intermediate visual goals, and predictive feedback generates the corresponding actions, improving cross-task mean success and reducing decision latency relative to AP (LeWM). For full-task execution, we adapt a standard imitation-learning policy head and evaluate on Push-T and LIBERO-Goal~\cite{liu2023libero}. Comparisons with recent imitation-learning, VLA, and WAM baselines show strong task performance and a favorable performance--computation trade-off. These results show that goal-aware predictive feedback effectively connects future prediction to action refinement across local goal reaching, hierarchical long-horizon control, and full-task execution.

\section{Related Work}
\label{sec:related}
\paragraph{JEPA and visual representation learning.}
JEPAs learn predictive representations in latent space without reconstructing pixels. This paradigm provides a foundation for world models that anticipate how the environment evolves under a given action sequence. I-JEPA predicts latent image-region representations~\cite{assran2023ijepa}, V-JEPA extends this approach to video~\cite{bardes2024vjepa}, and V-JEPA~2 connects video pretraining with action-conditioned prediction and robotic planning~\cite{assran2025vjepa2}. LeWM combines predictive learning with distributional regularization to learn visual encoders and action-conditioned latent dynamics from pixels~\cite{maes2026lewm}. Alternatively, DINOv2 provides transferable pretrained visual features~\cite{oquab2023dinov2}, and DINO-WM learns action-conditioned prediction over their dense spatial structure, enabling visual goal reaching through candidate-action search~\cite{zhou2024dinowm}. \method{} uses frozen DINOv2 with learned scene pooling and task-conditioned fusion to represent observations, predicted scenes, and goals in a common feature space. We feed predictions and goal discrepancies back into action generation for refinement.

\paragraph{Imitation learning models and vision--language--action.}
Imitation learning trains policies on demonstrations: ACT predicts action chunks~\cite{zhao2023act}, Diffusion Policy models action distributions through denoising~\cite{chi2023diffusion}, and VQ-BeT learns discrete action representations~\cite{lee2024vqbet}. Patch Policy connects dense pretrained visual features to lightweight action generators~\cite{zhou2026patchpolicy}. Many VLA models retain demonstration-based action learning while incorporating vision--language pretraining to connect visual observations and language instructions to control. OpenVLA-OFT combines parallel decoding, action chunking, and continuous action regression for efficient adaptation~\cite{kim2025openvlaoft}. VLA models also incorporate future prediction: DreamVLA forecasts dynamic, spatial, and semantic information to condition a diffusion action model~\cite{zhang2025dreamvla}. \method{} focuses on visual-goal-aware action refinement in latent space, explicitly using future predictions to guide task execution.

\begin{figure*}[t]
\centering
\includegraphics[width=0.95\textwidth,trim=0 5bp 0 12bp,clip]{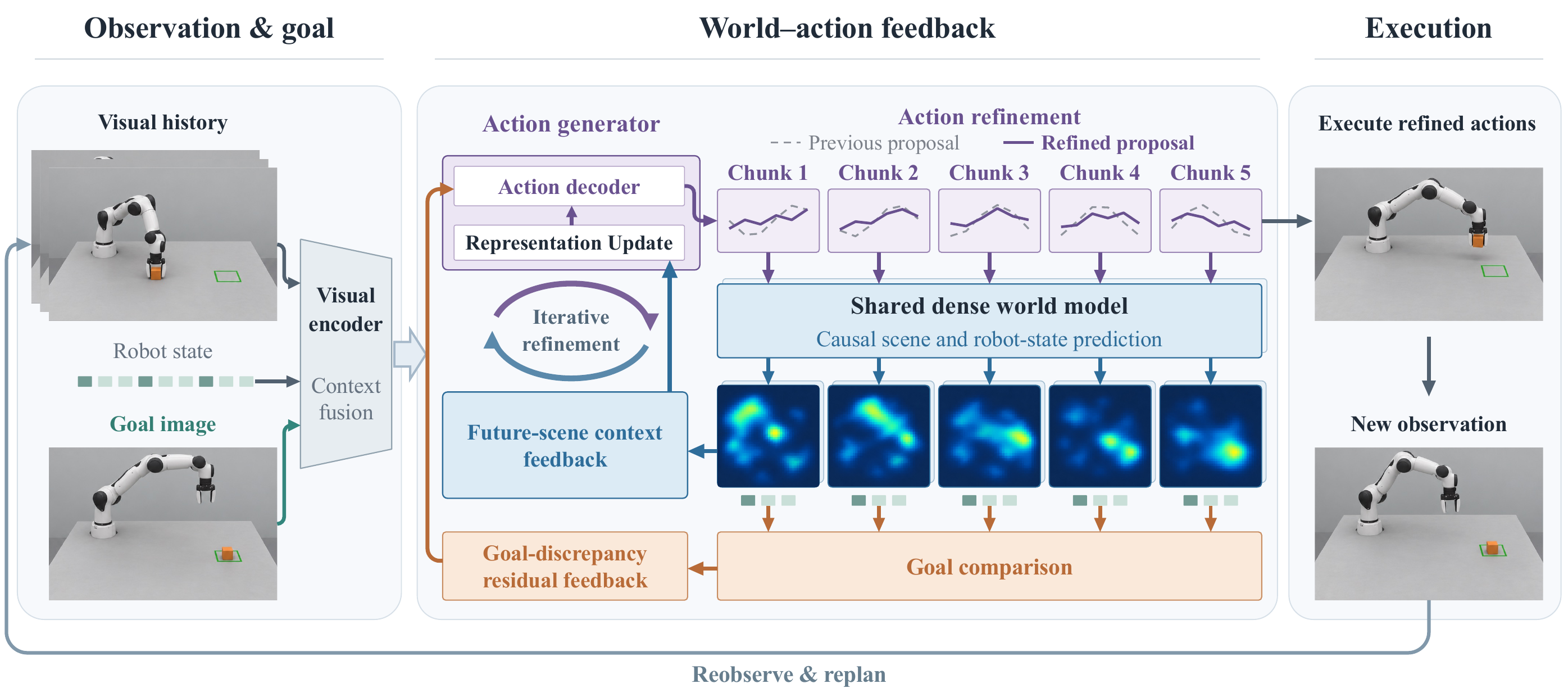}
\caption{\textbf{Overview of WA-JEPA.} The model iteratively refines an action plan before execution through a proposal--prediction--goal-aware feedback loop. Goal-discrepancy residual feedback guides chunk-wise action corrections, while future-scene context feedback coordinates updates across the plan. Both paths refine the same pending plan; new observations inform subsequent decisions as the process repeats until task completion.}
\label{fig:architecture}
\end{figure*}

\paragraph{World models and world--action models.}
\looseness=-1
World models support control through planning or policy learning. TD-MPC2 combines latent dynamics with value-based trajectory optimization~\cite{hansen2023tdmpc2}, and Dreamer learns policies through latent imagination~\cite{hafner2019dreamer}. WAMs bring future prediction and action generation into a unified model. DreamZero jointly generates video and actions using a pretrained video diffusion backbone~\cite{ye2026dreamzero}. RynnVLA-002 jointly learns action-conditioned image prediction and observation-conditioned action generation~\cite{cen2025rynnvla}. Fast-WAM retains video co-training but skips future prediction at inference to reduce decision latency~\cite{yuan2026fastwam}. LaDi-WM conditions a diffusion policy on predicted future features to iteratively refine its action outputs~\cite{huang2025ladiwm}. These designs differ in how predictions support action learning and inference. \method{} uses visual-goal-aware predictive feedback during inference: explicit discrepancies between predicted scenes and visual goals guide chunk-wise residual updates, while unpooled future-scene context coordinates refinement across the plan. Both paths update the same proposal before execution.

\section{World--Action Modeling through Predictive Feedback}
\label{sec:method}
\subsection{Problem formulation and overview}
We study robot control learned from offline demonstrations, where observations and task goals condition actions for completing specified tasks. At decision time $t$, the model receives the $L$ most recent images and corresponding proprioceptive states, $\{o_\tau,p_\tau\}_{\tau=t-L+1}^{t}$, together with an image $o_g$ depicting the task goal.
An action sequence $A\in\mathbb R^{H\times bd_a}$ consists of $H$ temporally ordered action chunks, each containing $b$ consecutive controls of dimension $d_a$.
As shown in Fig.~\ref{fig:architecture}, \method{} couples action generation with latent world prediction through goal-aware feedback. The model first generates an action proposal. The world model predicts scene features and proprioceptive states at each chunk boundary, and these predictions guide proposal refinement relative to the task goal. We use $h$ to index action chunks and $r$ to index internal refinement stages. Refinement updates the same pending plan without advancing environment time; new observations obtained after execution inform the next decision. We omit the batch dimension throughout.

\subsection{Representing the scene relative to the goal}
Predictive feedback requires future scenes and task goals to be compared in a common feature space. We use frozen DINOv2 to extract visual features from observation images, goal images, and future images used during training. Learnable multilayer feature fusion and query pooling summarize these features into scene tokens $Z_t\in\mathbb R^{S\times d}$, where $S$ is the number of scene tokens and $d$ is the feature dimension. The same fusion and pooling parameters are used for all images, so the world model's prediction targets and the goal encoding share a representation space. The scene tokens retain multiple learned summaries of the visual input; averaging them later provides a compact comparison, while retaining them separately supplies more detailed context for refinement.

We embed standardized proprioceptive states as an additional token and fuse it with each frame's scene features. Temporal attention then integrates the observation history. The resulting context combines historical visual information with proprioception, providing temporal context for both action generation and future prediction.

Bidirectional cross-attention exchanges information between the latest history-conditioned frame representation and the goal encoding $G$, producing context memory $M\in\mathbb R^{L(S+1)\times d}$ and goal memory $\widetilde G\in\mathbb R^{S\times d}$ for subsequent action generation. We also retain the current scene encoding $Z_0\in\mathbb R^{S\times d}$ before this interaction and the original goal encoding $G\in\mathbb R^{S\times d}$. The former provides the starting point for world-model prediction; the latter provides the reference for computing feature discrepancies between predicted scenes and the task goal. Module architectures and dimensions are detailed in App.~\ref{app:method_details}.

\subsection{Goal-aware dual-path predictive feedback}
At stage $r$, action tokens $Q^{(r)}\in\mathbb R^{H\times d}$ represent the plan, with one token per chunk. \emph{Goal-discrepancy residual feedback (GDRF)} updates these tokens, while \emph{future-scene context feedback (FSCF)} augments the memory read during the next proposal update. Both paths use the same predicted future, with observations and the goal fixed throughout refinement.

\paragraph{Generating an action proposal.}
The initial action representation $Q^{(1)}$ combines learned chunk queries and temporal embeddings with a conditioning signal derived from the pooled current scene $Z_0$ and its feature difference from the pooled goal $G$. At stage $r$, the action-update module $T_r$ first exchanges information among action chunks through self-attention. It then reads context memory $M$, goal memory $\widetilde G$, and the preceding stage's future-scene summaries $\widehat Z^{(r-1)}$ through cross-attention. FSCF supplies the preceding proposal's predictions as this additional context. The shared decoder $D$ converts the updated representation $\widetilde Q^{(r)}$ into controls, and the range mapping $\mathcal B$ produces the action proposal:
\begin{equation}
\begin{aligned}
\widetilde Q^{(r)}&=T_r\!\left(Q^{(r)},[M,\widetilde G,\widehat Z^{(r-1)}]\right),\\
A^{(r)}&=\mathcal B\!\left(D(\operatorname{LN}(\widetilde Q^{(r)}))\right).
\end{aligned}
\label{eq:action_proposal}
\end{equation}
Here, $\operatorname{LN}$ denotes layer normalization, $\mathcal B$ maps decoder outputs to the valid action range, and $[\cdot]$ denotes concatenation along the token sequence dimension, with predicted scenes represented as a sequence of scene tokens. We initialize $\widehat Z^{(0)}$ by repeating $Z_0$ along the future-time axis; subsequent stages use the preceding proposal's predicted scenes.

\paragraph{Predicting the proposal's consequences.}
The shared world model $W_\psi$ receives context memory $M$, the current scene encoding $Z_0$, and the complete action proposal $A^{(r)}$. It embeds the proposed controls and updates future-scene queries initialized from $Z_0$ with time and scene-token embeddings. These queries read the observed context and the proposed actions to predict scene features $\widehat Z^{(r)}$ and proprioceptive states $\widehat P^{(r)}$ at each action-chunk boundary:
\begin{equation}
(\widehat Z^{(r)},\widehat P^{(r)})=W_\psi(M,Z_0,A^{(r)}).
\label{eq:world_prediction}
\end{equation}
Here, $\psi$ denotes the world-model parameters, $\widehat Z^{(r)}\in\mathbb R^{H\times S\times d}$, and $\widehat P^{(r)}\in\mathbb R^{H\times d_p}$, where $d_p$ is the proprioceptive-state dimension. Temporal masks restrict prediction $h$ to the first $h$ action chunks and future queries no later than $h$. Each prediction therefore captures the accumulated consequences of executing the corresponding action prefix from the current state, while predictions across future times can be computed in parallel. Scene prediction operates directly in feature space, without reconstructing future images or requiring another visual-encoder pass.

\paragraph{Refining action representations through goal discrepancy.}
GDRF learns a residual update for each action chunk from its predicted outcome and the task goal. A visual feature discrepancy alone does not specify a control correction: the appropriate update also depends on the current proposal and the state it is predicted to produce. At feedback stage $r$, we average the $S$ scene tokens of prediction $h$ to obtain $\overline{\widehat Z}_h^{(r)}$, then subtract it from the pooled goal encoding $\overline G$. This goal discrepancy, the predicted scene, the action token $\widetilde Q_h^{(r)}$, and encoded proprioceptive prediction $E_p(\widehat P_h^{(r)})$ form the feedback input:
\begin{equation}
\begin{aligned}
U_h^{(r)}&=[\widetilde Q_h^{(r)},\overline{\widehat Z}_h^{(r)},\overline G-\overline{\widehat Z}_h^{(r)},E_p(\widehat P_h^{(r)})],\\
Q_h^{(r+1)}&=\widetilde Q_h^{(r)}+\sigma(\gamma_r)F_r(U_h^{(r)}).
\end{aligned}
\label{eq:action_feedback}
\end{equation}
Here, $\overline{(\cdot)}$ denotes averaging over scene tokens, and $[\cdot]$ denotes concatenation along the feature dimension. The encoder $E_p$ maps predicted proprioception to a $d$-dimensional feature space, giving $U_h^{(r)}\in\mathbb R^{4d}$. The feedback function $F_r$ maps the goal discrepancy together with the action token, predicted scene, and encoded proprioception to an action-token correction. A learned gate $\sigma(\gamma_r)$, where $\sigma$ is the sigmoid function, controls the correction magnitude. Within each stage, all action chunks share $F_r$ and are compared with the same task goal $\overline G$.

The residual updates the latent action representation, preserving the current proposal as the starting point for the next stage. The shared decoder $D$ decodes proposals at each stage for world-model prediction; only the final-stage action sequence is used for execution after refinement. Chunk-wise corrections are coupled: each predicted scene $\widehat Z_h^{(r)}$ reflects the accumulated consequences of the first $h$ action chunks, while self-attention in the next-stage module $T_{r+1}$ further coordinates updates across the plan.

FSCF complements GDRF by retaining details in individual predicted scene tokens. At the next stage, action tokens read the current proposal's predicted scenes through cross-attention, allowing refinement to use scene details compressed by mean pooling. GDRF provides residual corrections based on goal discrepancies, while FSCF supplies scene context describing the anticipated action outcomes. Together, the two paths produce the refined proposal $A^{(r+1)}$, whose predicted consequences guide the next refinement.

\begin{figure*}[!t]
\begin{minipage}{\textwidth}
\centering
\includegraphics[width=0.9\textwidth,trim=0 2bp 0 4bp,clip]{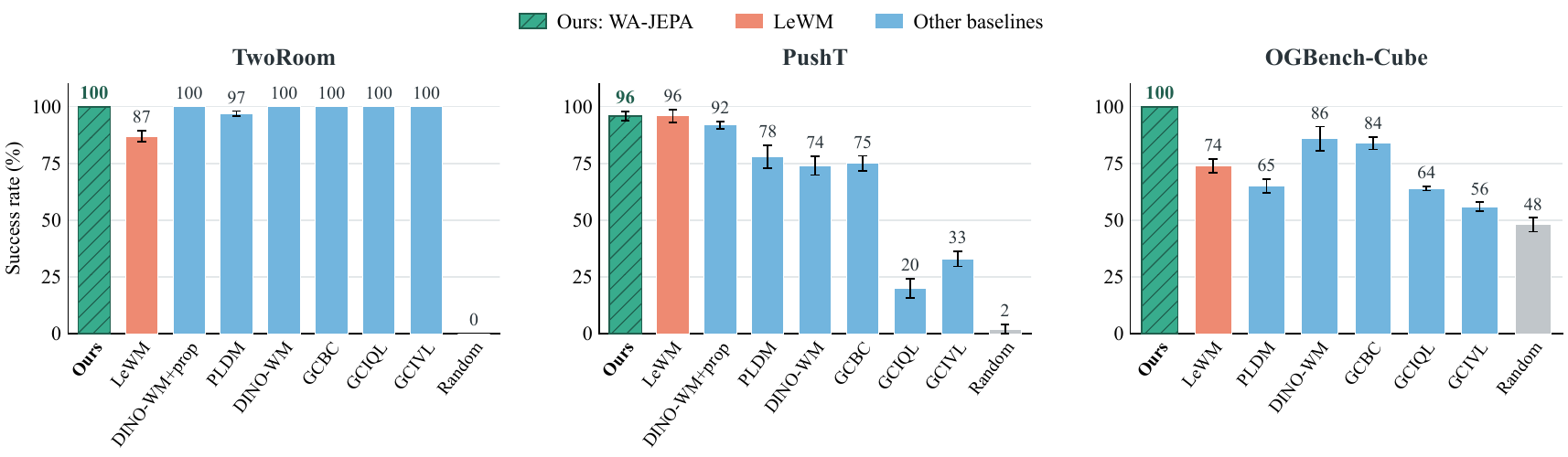}
\caption{Visual goal-reaching success on TwoRoom, PushT, and Cube. \method{} achieves 100\%, 96\%, and 100\%, respectively.}
\label{fig:mainperformance}
\end{minipage}
\end{figure*}

\begin{figure*}[!t]
\begin{minipage}{\textwidth}
\centering
\includegraphics[width=0.9\textwidth,trim=0 6bp 0 5bp,clip]{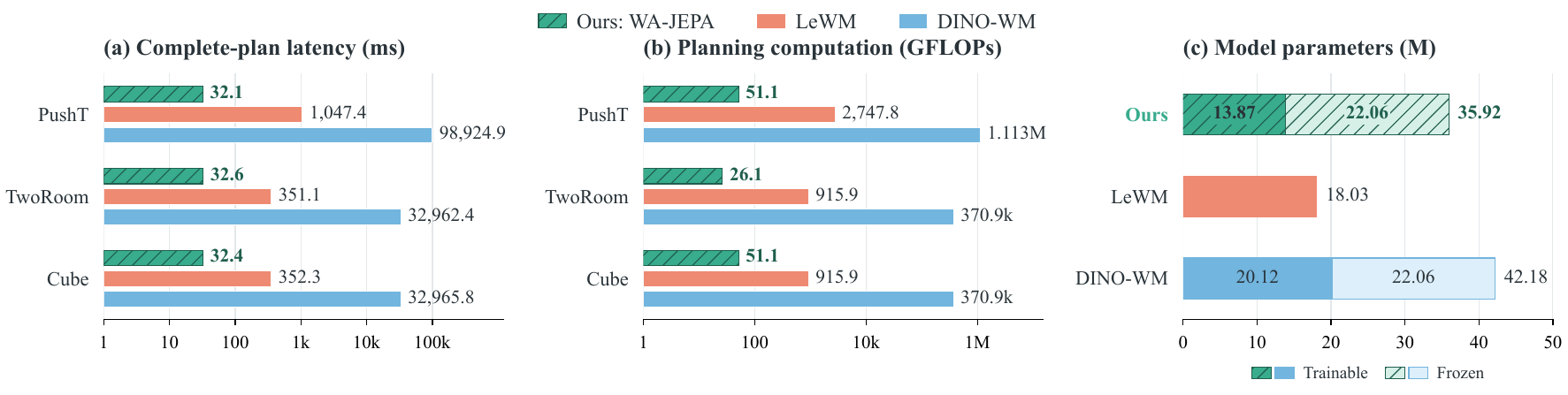}
\caption{\textbf{Short-horizon computational cost on an NVIDIA A40.} (a) Mean complete-plan latency (ms) and (b) neural arithmetic (GFLOPs). Both panels use logarithmic axes. (c) Total PushT model parameters (millions) on a linear axis, split by task-training status.}
\label{fig:efficiency}
\end{minipage}
\end{figure*}

\subsection{Jointly learning prediction and refinement}
\label{sec:loss}
We jointly learn action-conditioned prediction and action refinement from demonstrations, using recorded actions $A^*$, future scene features $Z^*$, and proprioceptive states $P^*$. The training objective combines prediction, action supervision, and outcome alignment:
\begin{equation}
\mathcal L_{\rm core}
=\mathcal L_{\rm pred}+\mathcal L_{\rm action}+\mathcal L_{\rm align}.
\label{eq:training_groups}
\end{equation}
The prediction loss $\mathcal L_{\rm pred}$ trains the world model to predict observed outcomes at each chunk boundary given recorded actions, using mean-squared error for scene features and SmoothL1 loss for proprioception. The action loss $\mathcal L_{\rm action}$ supervises intermediate and final proposals with demonstrated actions. For the final action proposal, $\mathcal L_{\rm align}$ penalizes discrepancies between the predicted scenes and proprioceptive states and their corresponding demonstration targets. It also encourages the predicted scene at the goal timestep to match the task-goal representation. Loss definitions and weights are provided in App.~\ref{app:method_details}.

\def\APTableRows{
LeWM (final goal) & 2.34 & 2.73 & $1.77\times$ & 1.56 & 1.56 & $2.07\times$ & 7.03 & 2.73 & $1.29\times$ & 3.65 & 2.34 & $1.71\times$\\
\method{} (final goal) & 4.69 & 3.52 & $0.33\times$ & 89.84 & 89.84 & $0.47\times$ & 35.16 & 19.14 & $0.70\times$ & 43.23 & 37.50 & $0.50\times$\\
Direct replay & 51.56 & 17.97 & $0.66\times$ & 99.22 & \textbf{100.00} & $0.53\times$ & 82.03 & 47.66 & $0.87\times$ & 77.60 & 55.21 & $0.69\times$\\
\midrule
AP (LeWM) & \textbf{69.53} & \textbf{64.84} & $1.99\times$ & 46.88 & 46.88 & $2.18\times$ & 46.88 & 29.69 & $1.33\times$ & 54.43 & 47.14 & $1.83\times$\\
AP (\method{}) & 56.25 & 46.48 & $1.00\times$ & \textbf{100.00} & \textbf{100.00} & $1.00\times$ & \textbf{84.38} & \textbf{52.34} & $1.00\times$ & \textbf{80.21} & \textbf{66.28} & $1.00\times$\\
}

\section{Experiments}
\label{sec:experiments}
We evaluate \method{} in three complementary settings: short-horizon planning, long-horizon control, and full-task execution. Short-horizon planning follows LeWM's goal-reaching protocol to compare control performance with world-model planners, imitation-learning methods, and offline reinforcement-learning methods. Long-horizon control combines model-driven action generation with high-level target selection to assess whether our method supports distant goals without retraining. Full-task execution evaluates planar pushing on Push-T and dual-view robot manipulation on LIBERO-Goal, using a VQ-BeT action head to compare against recent imitation-learning policies, vision language action (VLA) methods, and world action models (WAMs). This setting examines predictive feedback across distinct observation and action spaces, together with its deployment cost. Ablations examine the predictive-feedback system, refinement depth, and the contributions of its two feedback paths.

\subsection{Short-horizon experiments}
\label{sec:success}
\label{sec:efficiency}

We evaluate PushT, TwoRoom, and OGBench-Cube following LeWM's goal-reaching protocol~\cite{maes2026lewm}. For each task, 50 recorded initial states are paired with goal images 25 steps later in the source trajectories. The model executes its own actions for up to 50 environment steps, replanning after each 25-action buffer and stopping early on task success. Success uses the benchmark's physical predicates, rather than visual-feature distance. Fig.~\ref{fig:mainperformance} compares against the world-model planners LeWM, DINO-WM, and PLDM~\cite{maes2026lewm,zhou2024dinowm,sobal2025pldm}, the imitation-learning method GCBC~\cite{ghosh2019gcbc}, and offline reinforcement-learning methods GCIQL and GCIVL~\cite{kostrikov2021iql,park2025ogbench}. Baseline success rates follow the results reported by LeWM; each method retains its training recipe.

\begin{table*}[!t]
\begin{minipage}{\textwidth}
\centering\small
\caption{Long-range success (\%) and relative decision time under AP. Timing pools both start conditions and reports decision-time ratios relative to AP (\method{}). Bold marks the best success among the listed variants.}
\label{tab:ap_success}
\setlength{\tabcolsep}{3.5pt}
\resizebox{0.9\textwidth}{!}{
\begin{tabular*}{\textwidth}{@{\extracolsep{\fill}}lrrrrrrrrrrrr@{}}
\toprule
& \multicolumn{3}{c}{PushT} & \multicolumn{3}{c}{TwoRoom} & \multicolumn{3}{c}{Cube} & \multicolumn{3}{c}{Mean}\\
\cmidrule(lr){2-4}\cmidrule(lr){5-7}\cmidrule(lr){8-10}\cmidrule(lr){11-13}
Method & Std. & Pert. & Time$\downarrow$ & Std. & Pert. & Time$\downarrow$ & Std. & Pert. & Time$\downarrow$ & Std. & Pert. & Time$\downarrow$\\
\midrule
\APTableRows
\bottomrule
\end{tabular*}
}
\end{minipage}
\end{table*}

\paragraph{Configuration.}
Our short-horizon experiments use a lightweight MLP action head to decode action representations into continuous control sequences. We use frozen DINOv2-Small features, eight scene tokens per frame, width $d=192$, and four proposal stages with three feedback updates. Each proposal contains five chunks of five controls; world predictions correspond to the ends of these chunks. History comprises three frames on PushT/Cube and one on TwoRoom. Task-specific models are trained with AdamW at peak learning rate $5\times10^{-5}$, using direct action and latent-prediction supervision.

\method{} matches LeWM's reported 96\% success on PushT and reaches 100\% on TwoRoom and Cube, attaining the highest or joint-highest reported success in all three tasks. The results show that learned proposal refinement can retain strong local control performance while replacing online candidate optimization with a fixed-depth computation.

\paragraph{Computational cost.}
We measure complete-plan latency and neural arithmetic on an NVIDIA A40 at batch size one, using selected \method{} checkpoints and publicly released checkpoints for both LeWM and DINO-WM. Timing includes preprocessing, transfers, and action generation, excluding loading and simulator execution. Both world-model planners use 300 candidates and 30 elites, with 30 CEM iterations on PushT and 10 on TwoRoom/Cube. Native deployment precision is retained: BF16 autocast for \method{} and DINO-WM, and FP32 for LeWM. Complete measurement details appear in App.~\ref{app:latency}.

Fig.~\ref{fig:efficiency} (a) shows a clear inference-speed advantage: \method{} is $10.8$--$32.6\times$ faster than LeWM, including 32.13\,ms versus 1,047.36\,ms on PushT. DINO-WM incurs substantially more computation under the measured search settings. Our model encodes the observations once and refines one proposal through shared world predictions, whereas CEM repeatedly evaluates a population of action sequences. Consistent with this difference, \method{} uses only 1.9--5.6\% of the FLOPs required by LeWM for planning, as shown in Fig.~\ref{fig:efficiency} (b).
Fig.~\ref{fig:efficiency} (c) compares parameter counts. Both \method{} and DINO-WM include frozen DINOv2 weights, while all LeWM parameters are trainable. \method{} has the fewest trainable parameters of the three models: 13.87\,M on PushT, compared with 18.03\,M for LeWM and 20.12\,M for DINO-WM. \method{} achieves strong control performance with low computational cost.
Search-budget sweeps and the encoding-cache comparison are provided in App.~\ref{app:cem_budget_followup}.

\subsection{Long-horizon experiments}
\label{sec:longgoal}

\begin{table*}[!t]
\begin{minipage}{\textwidth}
\centering
\caption{\textbf{Full-task performance and computational cost.} Scores are percentages (mean $\pm$ available SD). Cost pairs denote \textbf{Push-T / LIBERO-Goal}; unpaired costs refer only to \textbf{LIBERO-Goal}. Memory is peak allocated GPU memory. Bold marks the best values within each task.}
\label{tab:fulltask_pusht}
\fontsize{9}{11}\selectfont
\setlength{\tabcolsep}{2.4pt}
\renewcommand{\arraystretch}{1.20}
\newcommand{\costpair}[3]{\makebox[#1][r]{#2}\kern0.16em/\kern0.16em\makebox[#1][r]{#3}}
\newcommand{\parampair}[2]{\makebox[2.75em][r]{#1}\kern0.16em/\kern0.16em\makebox[2.75em][r]{#2}}
\newcommand{\memorypair}[2]{\costpair{3.25em}{#1}{#2}}
\newcommand{\floppair}[2]{\costpair{2.75em}{#1}{#2}}
\newcommand{\latencypair}[2]{\costpair{2.75em}{#1}{#2}}
\resizebox{0.9\textwidth}{!}{
\begin{tabular*}{\textwidth}{@{\extracolsep{\fill}}llccrrrr@{}}
\toprule
\textbf{Method} & \textbf{Head} & \textbf{Push-T} & \textbf{LIBERO-Goal} & \multicolumn{1}{c}{\textbf{Params.} (M)$\downarrow$} & \multicolumn{1}{c}{\textbf{Mem.} (MiB)$\downarrow$} & \multicolumn{1}{c}{\textbf{FLOPs} (G)$\downarrow$} & \multicolumn{1}{c}{\textbf{Latency} (ms)$\downarrow$} \\
\midrule
WebSSL (CLS)~\cite{zhou2026patchpolicy} & VQ-BeT & $59\pm1$ & $95\pm1$ & \parampair{332.7}{308.0} & \memorypair{665.0}{647.3} & \floppair{162.5}{324.6} & \latencypair{19.20}{17.49} \\
Patch Policy~\cite{zhou2026patchpolicy} & VQ-BeT & $68\pm3$ & $94\pm1$ & \parampair{334.0}{308.7} & \memorypair{894.5}{2552.6} & \floppair{255.2}{413.5} & \latencypair{26.49}{81.35} \\
ResNet-18 Patch~\cite{zhou2026patchpolicy} & ACT & $64\pm3$ & $93\pm2$ & \parampair{66.5}{66.7} & \memorypair{144.0}{152.6} & \floppair{\textbf{6.1}}{\textbf{11.9}} & \latencypair{\textbf{11.87}}{\textbf{12.01}} \\
OpenVLA-OFT~\cite{kim2025openvlaoft} & MLP & $59\pm2$ & 95 & \parampair{7608.4}{7709.2} & \memorypair{14837.7}{15331.6} & \floppair{4202.6}{8766.6} & \latencypair{91.13}{147.16} \\
Fast-WAM~\cite{yuan2026fastwam} & Flow matching & --- & 96 & 12406.3 & 23959.7 & 2545.1 & 438.53 \\
RynnVLA-002~\cite{cen2025rynnvla} & Regression & --- & 96 & 7150.3 & 15429.3 & 16297.3 & 436.55 \\
DreamVLA~\cite{zhang2025dreamvla} & Diffusion & --- & 86 & 734.0 & 1515.7 & 1001.8 & 85.77 \\
\midrule
\textbf{WA-JEPA} & VQ-BeT & $\mathbf{71\pm2}$ & $\mathbf{97\pm1}$ & \parampair{\textbf{37.2}}{\textbf{37.8}} & \memorypair{\textbf{92.2}}{\textbf{115.6}} & \floppair{13.5}{26.5} & \latencypair{27.21}{26.95} \\
\bottomrule
\end{tabular*}
}
\end{minipage}
\end{table*}

To assess whether our method supports distant goals, we integrate \method{} into Anchored Planning (AP), the long-horizon planning framework developed for LeWM in \emph{Aim Short to Reach Far}~\cite{liu2026aimshort}, on PushT, TwoRoom, and Cube. AP matches the current observation and final goal to a segment in an offline trajectory memory and selects an early successor observation as an intermediate visual goal, turning distant-goal reaching into a sequence of local decisions. Within this hierarchical framework, AP retrieves intermediate visual goals, while \method{} generates actions toward them through predictive feedback, without candidate-action search or retraining.

We evaluate 128 standard starts from recorded query states and 256 perturbed starts obtained by independently applying two predefined five-action prefixes to each state. Final goals lie 140, 100, and 150 steps ahead on PushT, TwoRoom, and Cube; execution budgets are 280, 200, and 150 steps, respectively. All methods share queries, success criteria, and a five-step decision cadence. We retrieve trajectories from the same memory bank to preserve the high-level goal-selection mechanism.

In Tab.~\ref{tab:ap_success}, Final conditions the corresponding model directly on the final task goal, without selecting intermediate goals. Direct retrieves a demonstration segment matched to the current observation and task goal and executes its recorded actions. AP instead selects intermediate visual goals from retrieved demonstrations and uses the corresponding model to generate actions toward them: AP (LeWM) optimizes action sequences through CEM, while AP (\method{}) generates and refines proposals through goal-aware predictive feedback.

Intermediate goal selection improves \method{}'s long-horizon goal-reaching performance, reflecting the complementary roles of high-level planning and local predictive refinement. Relative to AP (LeWM), AP (\method{}) improves cross-task mean success by an absolute 25.78\% and 19.14\% under standard and perturbed starts, respectively, and outperforms direct replay under both conditions.
AP (\method{}) also achieves a $1.33$--$2.18\times$ speedup in mean decision time over AP (LeWM). Unlike the short-horizon timing of complete 25-action plans, AP timing includes target retrieval and rendering shared by both methods. AP (LeWM) also searches only a five-action chunk using cached visual encodings, while \method{} still generates 25 actions and executes the first five. The cheaper baseline search and shared overhead reduce the overall speedup in this hierarchical setting. Nevertheless, \method{} retains an online decision-speed advantage while achieving higher cross-task mean success.
\subsection{Full-task evaluation}
\label{sec:fulltask_pusht}

We evaluate full-task performance and deployment efficiency on Push-T and LIBERO-Goal~\cite{liu2023libero} against recent imitation-learning, VLA, and WAM methods~\cite{zhou2026patchpolicy,zhao2023act,kim2025openvlaoft,yuan2026fastwam,cen2025rynnvla,zhang2025dreamvla}. For these tasks, we adapt \method{} with a VQ-BeT action head. The adapted models retain action-conditioned future prediction and both feedback paths, iteratively refining action proposals using predicted outcomes and task goals before execution. Observation inputs, action horizons, and execution settings follow the respective task configurations, as detailed in App.~\ref{app:fulltask_protocol}.

Tab.~\ref{tab:fulltask_pusht} reports final coverage for Push-T and task success rate for LIBERO-Goal, both as percentages. For several imitation-learning baselines, we use the results reported in Patch Policy~\cite{zhou2026patchpolicy}. For Fast-WAM~\cite{yuan2026fastwam} and DreamVLA~\cite{zhang2025dreamvla}, we re-evaluate their public checkpoints on the same LIBERO-Goal cases and under the same environment protocol as \method{}. We measure computational cost on an NVIDIA A40 with BF16 and batch size 1. Latency measures one complete online planning call that generates an action chunk. Our task scores report the mean and standard deviation across three random-seed runs, each independently evaluated once.

\begin{figure*}[!t]
\begin{minipage}{\textwidth}
\centering
\includegraphics[width=0.9\textwidth]{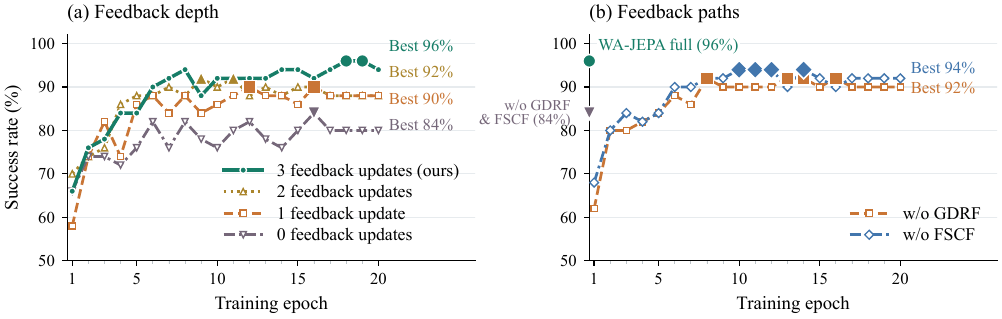}
\caption{\textbf{Predictive-feedback ablations on PushT.} (a) Zero to three feedback updates. (b) Variants without GDRF or FSCF. Curves show 20 unsmoothed checkpoints from one training seed; enlarged markers and Best labels indicate peak success.}
\label{fig:ablations}
\end{minipage}
\end{figure*}

\begin{figure*}[!t]
\begin{minipage}{\textwidth}
\centering
\includegraphics[width=0.9\textwidth,trim=0 15bp 0 14bp,clip]{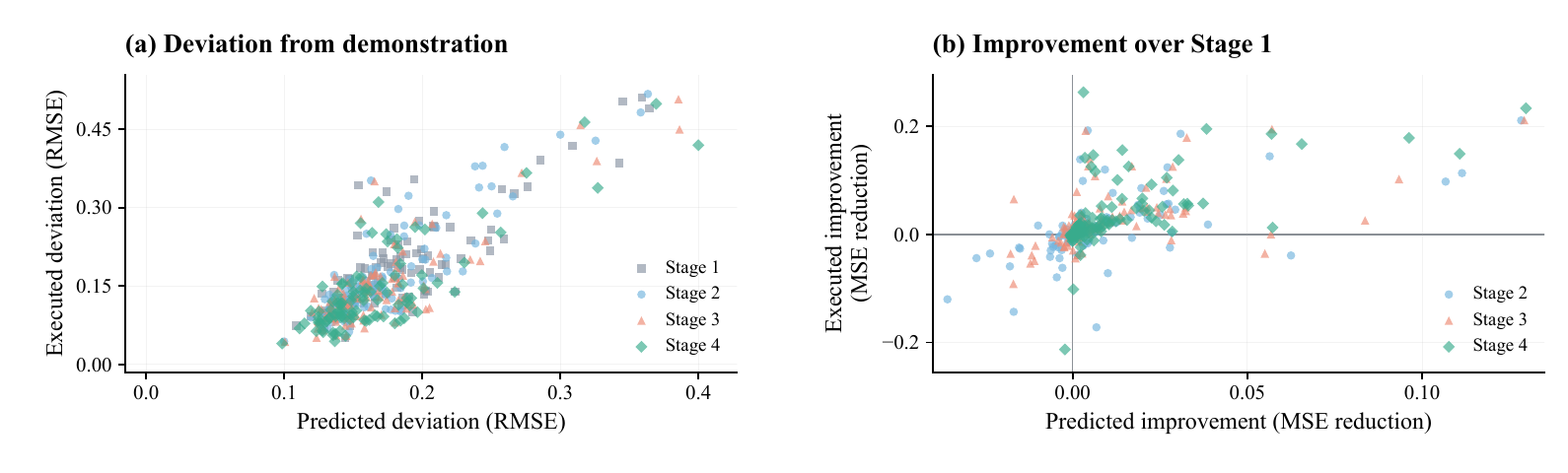}
\caption{\textbf{Prediction--execution correspondence on PushT.} (a) Predicted and executed scene deviations from the outcomes of demonstrated actions, measured by feature RMSE across chunk boundaries for proposals from stages 1--4. (b) Reduction in terminal goal-feature MSE relative to stage~1, comparing predicted and executed improvements for stages 2--4; positive values indicate improvement. Each point represents one stage proposal in one case.}
\label{fig:prediction_execution}
\end{minipage}
\end{figure*}

\method{} achieves $71\pm2$\% final coverage on Push-T and $97\pm1$\% task success on LIBERO-Goal, with both means exceeding the available baseline results for the corresponding task in Tab.~\ref{tab:fulltask_pusht}. These comparisons include imitation-learning and VLA policies that directly generate actions, as well as WAMs. Together, these results demonstrate the effectiveness of WA-JEPA in pushing and robot manipulation tasks. By using action-conditioned future predictions to guide goal-aware action refinement, WA-JEPA enables the model to combine perceived task goals with its learned understanding of physical dynamics to improve task execution.

\paragraph{Computational cost.}
Compared with Patch Policy, \method{} reduces FLOPs per online planning call by 94.7\% on Push-T and 93.6\% on LIBERO-Goal, while using fewer parameters and less peak GPU memory on both tasks. Decision latency is comparable to Patch Policy on Push-T and lower on LIBERO-Goal. On LIBERO-Goal, \method{} also outperforms the listed VLA and WAM baselines in task success, with lower decision latency and smaller computational and memory requirements. Together, these results demonstrate a favorable trade-off between task performance and deployment efficiency, while retaining explicit future prediction and iterative action refinement.
\subsection{Ablation studies}
\label{sec:ablations}
We examine feedback depth and the two feedback paths on PushT in the short-horizon goal-reaching setting, with the same experimental conditions across variants. Depth variants use zero to three updates. Path variants retain four proposal stages, the world model, and all training objectives, disabling either goal-discrepancy residual feedback (GDRF) or future-scene context feedback (FSCF).

Fig.~\ref{fig:ablations} (a) shows that the full model achieves 96\% best-checkpoint success, compared with 84\% without predictive feedback. Additional updates improve best-checkpoint success among feedback-enabled variants, supporting successive refinement using predicted consequences. The zero-update variant retains direct action learning while removing world prediction and its associated feedback components and objectives; this comparison therefore measures the overall benefit of the predictive-feedback system.

In Fig.~\ref{fig:ablations} (b), we use the same marker conventions as in panel (a) to indicate the best success rates of the full model and the zero-feedback system variant without either feedback path. Both reference scores are taken from panel (a). Removing either GDRF or FSCF reduces peak success performance, indicating that both feedback paths contribute to action refinement. GDRF uses discrepancies between predicted scenes and the task goal to guide chunk-wise corrections, while FSCF supplies future-scene context to coordinate adjustments across the plan. Combining both paths yields better control performance.

\paragraph{Prediction--execution correspondence.}
We use a fixed checkpoint on 100 newly sampled PushT validation cases to examine the correspondence between predicted outcomes and actual execution. For each case, we obtain proposals from all four stages and execute each proposal and the demonstrated action sequence independently. Before each rollout, we restore the same initial state and keep the task goal fixed. Each rollout lasts 25 steps without replanning or early stopping.

Fig.~\ref{fig:prediction_execution} (a) uses the scenes obtained by executing demonstrated actions as a common reference and compares predicted and executed deviations for each stage proposal. The two deviations are positively associated: proposals predicted to produce scenes closer to the demonstration outcomes generally do so in actual execution. Pooling stages 1--4 across the 100 cases gives a Pearson correlation of $r=0.844$. This correspondence indicates that the model can estimate how far a proposal's outcome deviates from the demonstration, providing a basis for evaluating proposals before execution.

Fig.~\ref{fig:prediction_execution} (b) further examines whether these predictions support effective action refinement. Relative to stage~1, reductions in predicted goal error are positively associated with reductions in executed goal error. Most stage~4 points lie in the first quadrant, indicating that the final refined proposal generally brings the actual scene closer to the task goal. Together, these results show correspondence between predicted and actual action consequences, with predicted improvements also reflected in execution, supporting goal-aware predictive feedback for action refinement.

\section{Conclusion}
\label{sec:conclusion}
We presented \method{}, a world--action model that uses goal-aware predictive feedback to refine action proposals before execution without candidate-action search. It connects learned dynamics with action generation by making predicted consequences part of the decision process. Experiments on short-horizon goal reaching, hierarchical long-horizon control, and full-task execution show strong task performance with low online computational cost. Full-task comparisons with recent imitation-learning, VLA, and WAM methods show higher task scores, with lower decision latency than the compared VLA and WAM baselines on LIBERO-Goal. These findings support goal-aware predictive feedback as a practical approach to efficient action generation across different control settings.

\label{maintext:end}
{\small\bibliographystyle{ieeenat_fullname}\bibliography{main}}
\clearpage
\appendix
\renewcommand{\theHequation}{supp.\thesection.\arabic{equation}}
\setcounter{table}{0}
\setcounter{figure}{0}
\renewcommand{\thetable}{S\arabic{table}}
\renewcommand{\thefigure}{S\arabic{figure}}
\renewcommand{\theHtable}{supp.\arabic{table}}
\renewcommand{\theHfigure}{supp.\arabic{figure}}
\noindent The supplementary material presents action refinement and goal-discrepancy ablations (App.~\ref{app:additional_evaluations}), computational efficiency and search-budget analysis (App.~\ref{app:latency}), and architecture, training, and evaluation details (App.~\ref{app:method_details}).
\section{Additional Ablation Experiments}
\label{app:additional_evaluations}
We first evaluate action refinement in short-horizon, long-horizon, and full-task settings, using one trained four-stage model per task. Each run executes the proposal $A^{(k+1)}$ produced after $k\in\{0,1,2,3\}$ feedback updates. The model always computes all four stages; only the executed proposal changes. These experiments complement the feedback-depth ablation in the main paper: that ablation trains separate models with different numbers of feedback updates, whereas here we evaluate proposals after different numbers of updates within the same fully trained model. They do not measure the latency of stopping inference early. A separate training ablation then compares explicit goal differences with direct goal concatenation.

\subsection{Fixed-network refinement ablation: short-horizon tasks}
\label{app:fixed_short_stages}
For each task, we evaluate all four proposals on the same 50 initial--goal pairs as in the short-horizon experiment. Each run starts from the same physical state and executes one plan for up to 25 steps, stopping on success without replanning. All feedback settings use the same initial observations and task goals, with random seeds fixed across runs.

\begin{table}[!htbp]
\centering
\caption{Short-horizon task performance of a fixed model with different numbers of feedback updates.}
\label{tab:fixed_short_stages}
\small\setlength{\tabcolsep}{2.5pt}\renewcommand{\arraystretch}{1.12}
\begin{tabular*}{\linewidth}{@{\extracolsep{\fill}}lrrr@{}}
\toprule
Updates & PushT & TwoRoom & Cube \\
\midrule
0 & 16 / 45.37 & 100 / 13.99 & 100 / 2.42 \\
1 & 52 / 32.53 & 100 / 14.14 & 96 / 2.47 \\
2 & 68 / 24.42 & 100 / 14.13 & 100 / 2.44 \\
3 & 96 / 18.30 & 100 / 14.08 & 100 / 2.41 \\
\bottomrule
\end{tabular*}
\par\smallskip{\footnotesize Each cell: success (\%) / position error. Position errors are in simulator pixels for PushT/TwoRoom and cm for Cube.\par}
\end{table}

Tab.~\ref{tab:fixed_short_stages} shows that feedback improves both goal attainment and final-state accuracy on PushT: success rises from 16\% to 96\%, while mean position error falls from 45.37 to 18.30. TwoRoom and Cube already achieve 100\% success with the initial proposal; their final proposals retain this level. Position error is the Euclidean distance between concatenated agent and object positions for PushT, agent--goal distance for TwoRoom, and object--goal distance for Cube.

\subsection{Fixed-network refinement ablation: long-horizon tasks}
\label{app:ap_fixed_stages}
We use the models, retrieval memory, queries, success criteria, and action budgets from App.~\ref{app:ap_protocol}. At each decision, AP selects an intermediate visual goal.
The evaluation comprises 4,608 rollouts, with 128 standard and 256 perturbed starts per task and stage. All stages use the same initial states, task goals, and evaluation settings.

\begin{table}[!htbp]
\centering
\caption{Fixed-checkpoint AP (WA-JEPA) success (\%) after 0--3 feedback updates. Mean averages the three tasks equally.}
\label{tab:ap_fixed_stages_interim}
\small\setlength{\tabcolsep}{3pt}\renewcommand{\arraystretch}{1.12}
\begin{tabular*}{\linewidth}{@{\extracolsep{\fill}}llrrrr@{}}
\toprule
Task & Start & \multicolumn{4}{c}{Feedback updates} \\
\cmidrule(lr){3-6}
 & & 0 & 1 & 2 & 3 \\
\midrule
PushT & Std. & 4.69 & 17.97 & 32.81 & \textbf{56.25} \\
 & Pert. & 5.86 & 16.02 & 28.52 & \textbf{46.48} \\
\addlinespace[2pt]
TwoRoom & Std. & 99.22 & \textbf{100.00} & \textbf{100.00} & \textbf{100.00} \\
 & Pert. & \textbf{100.00} & \textbf{100.00} & \textbf{100.00} & \textbf{100.00} \\
\addlinespace[2pt]
Cube & Std. & 79.69 & 80.47 & 82.03 & \textbf{84.38} \\
 & Pert. & 46.09 & 50.78 & \textbf{53.91} & 52.34 \\
\midrule
Mean & Std. & 61.20 & 66.15 & 71.61 & \textbf{80.21} \\
 & Pert. & 50.65 & 55.60 & 60.81 & \textbf{66.28} \\
\bottomrule
\end{tabular*}
\end{table}

Tab.~\ref{tab:ap_fixed_stages_interim} shows that cross-task mean success increases with each feedback update, from 61.20\% to 80.21\% under standard starts and from 50.65\% to 66.28\% under perturbed starts.

\subsection{Fixed-network refinement ablation: full tasks}
\label{app:fixed_fulltask_stages}
For full-task evaluation on Push-T and LIBERO-Goal, we compare action proposals from different refinement stages using a fixed checkpoint per task under identical evaluation conditions, always computing all four stages before selecting the designated output.

\begin{table}[!htbp]
\centering
\caption{Full-task performance of a fixed model with different numbers of feedback updates. Each setting uses 100 cases per task; all values are percentages.}
\label{tab:fixed_fulltask_stages}
\small\setlength{\tabcolsep}{3pt}\renewcommand{\arraystretch}{1.12}
\begin{tabular*}{\linewidth}{@{\extracolsep{\fill}}lrrrr@{}}
\toprule
Task / feedback updates & 0 & 1 & 2 & 3 \\
\midrule
Push-T coverage & 31.00 & 60.30 & 59.65 & 70.78 \\
LIBERO-Goal success & 87 & 88 & 94 & 97 \\
\bottomrule
\end{tabular*}
\end{table}

Predictive refinement also improves complete-task execution (Tab.~\ref{tab:fixed_fulltask_stages}). From the initial to the final proposal, Push-T final coverage increases from 31.00\% to 70.78\%, and LIBERO-Goal success increases from 87\% to 97\%. These gains extend the benefit of refinement beyond short-horizon goal reaching.

\subsection{Goal discrepancy versus concatenation}
\label{app:goal_input_ablation}
Eq.~\ref{eq:action_feedback} supplies the feedback network with both a predicted scene and its discrepancy from the goal. For clarity, let $z=\overline{\widehat Z}_h^{(r)}$ and $g=\overline G$. The pair $[z,g-z]$ preserves the same information as $[z,g]$. The distinction is how this information is presented: $z$ describes the predicted state, while $g-z$ explicitly expresses its deviation from the goal. We therefore test whether this explicit relative representation benefits feedback learning compared with direct concatenation.

In the Goal-Concat variant, we replace $\overline G-\overline{\widehat Z}_h^{(r)}$ with $\overline G$ in the input to the GDRF feedback MLP $F_r$. The action token, pooled predicted scene, and encoded predicted proprioception remain in the input, whose dimension stays $4d$. Both variants therefore retain the same goal and predicted-scene information, parameter count, and feedback modules. FSCF, the residual update and its gate, and initial goal conditioning are unchanged.
Both variants use the same random seeds and evaluation conditions as the main PushT experiment.

\begin{table}[!htbp]
\centering
\caption{Goal-input ablation on PushT. Success rates are percentages over 50 fixed cases. Best and Final refer to the best of 20 checkpoints and epoch 20; Last 5 averages epochs 16--20 within the same training run.}
\label{tab:goal_input_ablation}
\small\setlength{\tabcolsep}{3pt}\renewcommand{\arraystretch}{1.12}
\begin{tabular*}{\linewidth}{@{\extracolsep{\fill}}lrrr@{}}
\toprule
Feedback input & Best & Final & Last 5 \\
\midrule
Goal difference (ours) & \textbf{96} & \textbf{94} & \textbf{94.4} \\
Goal-Concat & 92 & 92 & 91.6 \\
\bottomrule
\end{tabular*}
\end{table}

As shown in Tab.~\ref{tab:goal_input_ablation}, explicit goal differences yield a best success rate of 96\%, compared with 92\% for direct concatenation. The advantage also holds at the final checkpoint and over the last five epochs. Under the matched training setup, this comparison favors the $[z,g-z]$ parameterization over $[z,g]$, despite their information equivalence. It supports explicitly exposing the goal discrepancy while retaining the predicted scene as context for the correction. This comparison uses one training seed; the last-five-epoch average describes late-training performance, not variation across independent runs.

\section{Computational Efficiency and Measurement Details}
\label{app:latency}
\setcounter{equation}{0}
\renewcommand{\theequation}{\thesection.\arabic{equation}}

We first report short-horizon planning costs and examine the success--latency trade-off under different CEM search budgets of LeWM. We then present decision costs for full-task execution. Each comparison uses the measurement protocol specified below.

\subsection{Short-horizon complete-plan cost}
Fig.~\ref{fig:efficiency}(a) reports mean complete-plan latency. Tab.~\ref{tab:latency_variability} gives the standard deviation of the same measurements in milliseconds, so that small variations remain visible.

\paragraph{Summary statistics.}
For the $n$ recorded times $t_i$, we report the arithmetic mean $\bar{t}$ and descriptive standard deviation
\begin{equation}
\sigma_t = \sqrt{\frac{1}{n}\sum_{i=1}^{n}(t_i-\bar{t})^2}.
\end{equation}
The standard deviation summarizes latency variation across timed inputs and repeated executions.

\begin{table}[!htbp]
\centering
\caption{Complete-plan latency underlying Fig.~\ref{fig:efficiency}(a). Mean and standard deviation (SD) are in milliseconds; WA-JEPA and LeWM use 20 timed calls over ten input pairs; DINO-WM retains ten recorded calls. All use three warmups.}
\label{tab:latency_variability}
\small
\setlength{\tabcolsep}{4pt}
\begin{tabular*}{\linewidth}{@{\extracolsep{\fill}}llrr@{}}
\toprule
Task & Method & Mean (ms) & SD (ms) \\
\midrule
PushT & \method{} & 32.13 & 0.36 \\
 & LeWM & 1047.36 & 2.65 \\
 & DINO-WM + CEM & 98924.91 & 34.01 \\
\midrule
TwoRoom & \method{} & 32.62 & 0.42 \\
 & LeWM & 351.06 & 1.54 \\
 & DINO-WM + CEM & 32962.39 & 22.51 \\
\midrule
Cube & \method{} & 32.35 & 0.20 \\
 & LeWM & 352.28 & 1.01 \\
 & DINO-WM + CEM & 32965.76 & 19.29 \\
\bottomrule
\end{tabular*}
\end{table}

\begin{table*}[!t]
\centering
\caption{Locally reevaluated success and latency under different LeWM CEM search budgets, compared with WA-JEPA. Latency measures one complete plan on an A40. Dashes denote untested settings.}
\label{tab:cem_budget_followup}
\small\setlength{\tabcolsep}{5pt}\renewcommand{\arraystretch}{1.12}
\begin{tabular}{@{}lrrrrrr@{}}
\toprule
Method / candidates / iterations & \multicolumn{2}{c}{PushT} & \multicolumn{2}{c}{TwoRoom} & \multicolumn{2}{c}{Cube} \\
\cmidrule(lr){2-3}\cmidrule(lr){4-5}\cmidrule(lr){6-7}
 & Success (\%) & Latency (ms) & Success (\%) & Latency (ms) & Success (\%) & Latency (ms) \\
\midrule
LeWM / 30 / 3 & 50 & 73.35 & 76 & 74.85 & 62 & 72.46 \\
LeWM / 100 / 3 & 58 & 73.32 & 82 & 76.93 & 72 & 79.44 \\
LeWM / 300 / 1 & 14 & 36.84 & 68 & 36.21 & 42 & 37.23 \\
LeWM / 300 / 3 & 58 & 105.12 & 84 & 106.83 & 68 & 105.91 \\
LeWM / 300 / 10 & 94 & 345.24 & 90 & 351.06 & 72 & 352.28 \\
LeWM / 300 / 30 & 94 & 1047.36 & -- & -- & -- & -- \\
\midrule
WA-JEPA & 96 & 32.13 & 100 & 32.62 & 100 & 32.35 \\
\bottomrule
\end{tabular}
\end{table*}

\paragraph{Measurement setup.}
We measure complete-plan latency on an NVIDIA A40 with batch size one, four CPU threads, and three warmups. WA-JEPA and LeWM are timed 20 times over the same ten input pairs; DINO-WM retains its ten recorded measurements under the same timing boundary. Timing excludes checkpoint loading, cold initialization, and simulator execution. \method{} and DINO-WM use BF16 autocast with FP32 weights; LeWM uses FP32 with TF32 disabled. Each method retains its native deployment configuration. Both LeWM and DINO-WM use publicly released checkpoints.

\paragraph{Planning latency.}
Following LeWM~\cite{maes2026lewm}, CEM uses 300 candidates and 30 elites, with 30 iterations on PushT and 10 on TwoRoom/Cube. Both search methods start from one real image. DINO-WM retains its three-step predictor context and caches the current and goal encodings across iterations. \method{} uses four proposal stages and four evaluations of the shared world model, without a candidate population.

\paragraph{Neural arithmetic and model size.}
FLOPs are measured along each method's inference path, counting two operations per multiply--accumulate. DINO-WM encodes the current and goal images once per plan, whereas LeWM's original implementation repeats encoding at each CEM iteration.

\method{} requires 51.05/26.14/51.06 GFLOPs per plan on PushT/TwoRoom/Cube, compared with 2747.75/915.92/915.93 GFLOPs for LeWM (Fig.~\ref{fig:efficiency}(b)). This is 1.9--5.6\% of LeWM's planning FLOPs. DINO-WM requires approximately 1113/371/371 TFLOPs because it predicts dense patches for the candidate population across search iterations.

Fig.~\ref{fig:efficiency}(c) separates all stored model parameters by whether they are updated during task training. On PushT, \method{} has 35.92M parameters: 13.87M trainable and 22.06M in frozen DINOv2-Small. LeWM has 18.03M parameters, all trainable, including its visual encoder and predictor. The profiled DINO-WM model has 42.18M parameters: 20.12M trainable and 22.06M in frozen DINOv2-Small. The DINO-WM count includes its saved action and dummy-proprioception embeddings. Frozen parameters are not updated during task training.
\subsection{Search budget and success--latency trade-off}
\label{app:cem_budget_followup}

We test whether a smaller CEM search budget can reduce LeWM's latency while retaining goal-reaching success. Following LeWM~\cite{maes2026lewm}, we use the same 50 initial--goal pairs per task, sampled from recorded trajectories with seed 42. Each rollout starts from the recorded initial state, targets the image 25 steps later, and allows up to 50 actions, with replanning after 25 actions and early termination on success. The released model weights are fixed throughout; we vary the search budget in two complementary sweeps.

\paragraph{Search depth: varying iterations.}
We fix the candidate population at 300 and evaluate 1, 3, and 10 CEM iterations on all three tasks, with an additional 30-iteration setting on PushT. This sweep measures how many rounds of candidate-distribution updates are needed to obtain effective actions.

\paragraph{Search width: varying candidates.}
We fix the number of CEM iterations at three and evaluate populations of 30, 100, and 300 candidate action sequences. This sweep measures the effect of evaluating more alternatives in each iteration. In both sweeps, the elite set contains 10\% of candidates, capped at 30. WA-JEPA uses the same fixed configuration as the reference for both sweeps.
Measurements use an A40, batch size one, four CPU threads, three warmups, and 20 timed calls.

We evaluate LeWM's officially released checkpoints in our local environment under different CEM search budgets, using the same repeated-encoding inference path as the main-text timing benchmark. Default-budget latencies and WA-JEPA latency measurements are shared with Fig.~\ref{fig:efficiency}.

Tab.~\ref{tab:cem_budget_followup} shows that smaller CEM budgets reduce latency at the cost of goal-reaching success. With 300 candidates on PushT, increasing the iteration count from one to ten raises success from 14\% to 94\%, while latency increases from 36.84 to 345.24\,ms. Increasing it further to 30 yields no additional success gain but raises latency to 1,047.36\,ms. WA-JEPA achieves 96\% success at 32.13\,ms and attains the highest success rate across the three evaluated tasks.
We also test caching the current and goal encodings across CEM iterations. This reduces LeWM's default-budget latency to 781.72, 267.74, and 267.91\,ms on PushT, TwoRoom, and Cube, respectively, while WA-JEPA retains a latency advantage.

\subsection{Full-task decision cost}
\label{app:fulltask_costs}

\paragraph{Timing boundary and hardware.}
All methods are measured sequentially on an NVIDIA A40 with batch size one, BF16 weights and autocast, four CPU threads, and TF32 disabled. Compilation and quantization are disabled. Inputs are $224\times224$, except for RynnVLA-002's native $256\times256$ images and FP32 tokenizer under BF16 autocast. Measurements use synthetic RGB inputs. We use 100 warmups and 1,000 CUDA-synchronized wall-clock samples per model and mode.

Each timed call processes the new observation, including preprocessing, device transfer, visual encoding, action generation, and output aggregation. Previous visual features and fixed goals or prompts are cached where supported by the inference procedure; DreamVLA re-encodes its visual history. Loading, episode initialization, and simulation are excluded. Push-T supplies one new camera view per decision; LIBERO-Goal supplies two. WA-JEPA generates five actions and returns one temporally ensembled action. Latency is reported per call, not divided by five. Tab.~\ref{tab:cost_horizons} lists the number of history frames and output actions used for each method and task in these measurements.

\begin{table}[!htbp]
\centering
\caption{History frames / output actions per cost-measurement call. Each task uses its own deployment configuration; costs are not normalized by output length.}
\label{tab:cost_horizons}
\small\renewcommand{\arraystretch}{1.12}\setlength{\tabcolsep}{5pt}
\begin{tabular*}{\linewidth}{@{\extracolsep{\fill}}lcc@{}}
\toprule
Method & Push-T & LIBERO-Goal \\
\midrule
WebSSL (CLS, VQ-BeT) & $5/5$ & $10/1$ \\
Patch Policy (VQ-BeT) & $5/5$ & $10/1$ \\
ResNet-18 Patch (ACT) & $1/5$ & $1/5$ \\
OpenVLA-OFT & $1/5$ & $1/8$ \\
Fast-WAM & --- & $1/32$ \\
RynnVLA-002 & --- & $2/5$ \\
DreamVLA & --- & $7/3$ \\
WA-JEPA (VQ-BeT) & $5/5$ & $5/5$ \\
\bottomrule
\end{tabular*}
\end{table}

\paragraph{Memory, parameters, and arithmetic.}
Parameter counts include frozen encoders and all deployment modules loaded into memory. Peak memory is measured by \texttt{torch.cuda.max\_memory\_allocated} and includes resident weights and caches. It does not measure reserved memory or device-wide usage.

We count convolution, matrix multiplication, and attention from their dense tensor shapes, with two FLOPs per multiply--accumulate. This includes new-image encoding and any history re-encoding required by the method. Causal attention is counted at the full matrix size. Normalization, activations, softmax, elementwise operations, sampling, and CPU postprocessing are excluded from GFLOPs but included in latency.

\section{Architecture, Training, and Evaluation}
\label{app:method_details}
\setcounter{equation}{0}
\renewcommand{\theequation}{\thesection.\arabic{equation}}
This section brings together the network architecture, training objectives, and evaluation protocols. We first describe the continuous-head model used for short-horizon goal reaching and AP, then specify the training configurations and evaluation settings, including the full-task VQ-BeT adaptations.

\subsection{Architecture and tensor shapes}
We encode $224\times224$ RGB images with frozen DINOv2 ViT-S/14, using its 384-dimensional patch features from layers 3, 7, and 11 (zero-indexed). Each layer's features are normalized, projected to $d=192$, and combined with learned softmax weights. Eight learned pooling queries produce the scene tokens. A two-layer MLP converts standardized proprioception into one token per frame. One attention layer fuses the nine tokens within each frame, and two further layers fuse the $9L$ tokens across history frames.

Tab.~\ref{tab:architecture_shapes} lists the tensor shapes for this configuration. We set $S=8$, $d=192$, and $H=b=5$, with $L=3$ for PushT/Cube and $L=1$ for TwoRoom.

\begin{table}[!htbp]
\centering
\caption{Core tensors for the continuous-head goal-reaching model. $B$: batch size; $L$: history frames; $H$: action chunks; $b$: controls per chunk. The full-task VQ-BeT configuration is specified separately in App.~\ref{app:fulltask_protocol}.}
\label{tab:architecture_shapes}
\small\renewcommand{\arraystretch}{1.12}\setlength{\tabcolsep}{3pt}
\begin{tabular*}{\linewidth}{@{\extracolsep{\fill}}ll@{}}
\toprule
Representation & Shape \\
\midrule
Current scene $Z_0$; goal $G,\widetilde G$ & $B\times8\times192$ \\
Context memory $M$ & $B\times9L\times192$ \\
Action tokens $Q^{(r)}$ & $B\times5\times192$ \\
Action proposal $A^{(r)}$ & $B\times5\times(5d_a)$ \\
Future scene $\widehat Z^{(r)}$ & $B\times5\times8\times192$ \\
Future state $\widehat P^{(r)}$ & $B\times5\times d_p$ \\
FSCF memory tokens & $B\times40\times192$ \\
GDRF input $U^{(r)}$ & $B\times5\times768$ \\
\bottomrule
\end{tabular*}
\end{table}

When training with three-frame histories, we retain one, two, or three real frames with probabilities 0.125, 0.125, and 0.75, respectively. Missing frames are filled by repetition and identified by validity embeddings. This augmentation is not used for TwoRoom.

The action embedding MLP has widths $bd_a\rightarrow768\rightarrow192$. Three world-update layers process the 40 future scene tokens. Four proposal stages provide three feedback updates. Each stage has its own action-update and feedback modules, while the action decoder and world model are shared across stages.

\paragraph{Action generation and FSCF.}
Each initial action token combines a learned chunk query, time and type embeddings, and features of the current scene and its difference from the goal. The conditioning network uses LayerNorm and a $2d\rightarrow d\rightarrow d$ MLP with GELU. Each update module $T_r$ applies self-attention, cross-attention, and a feed-forward layer with residual connections; attention inputs are layer-normalized. The shared decoder $D$ is a $d\rightarrow2d\rightarrow bd_a$ MLP with GELU. In Eq.~\ref{eq:action_proposal}, $\operatorname{LN}$ normalizes its input, and $\mathcal B$ applies $\tanh$ and rescales the output to the allowed action range.

FSCF adds learned time embeddings to the predicted scene tokens and arranges them into a sequence of $HS$ tokens. The next stage reads this sequence through cross-attention, retaining information from every predicted time and scene slot. This cross-attention uses no temporal causal mask because the current proposal's predictions are available before execution. Each stage decodes all action chunks together. Iteration occurs between proposal stages, not sequentially along the action timeline.

\paragraph{World-prediction modules.}
The action encoder maps each chunk to $d$ features, with SiLU between its linear layers. Future queries start from the sum of $Z_0$, learned queries, and time and slot embeddings. Three layers update them using causal self-attention and cross-attention to the context and actions. The $h$-th prediction attends only to future queries and action chunks up to $h$. Each layer also combines the future query, corresponding action features, and current scene through a gated residual update. The scene head uses LayerNorm and a $d\rightarrow2d\rightarrow d$ MLP. The proprioception head concatenates pooled predicted-scene features with action features, then uses LayerNorm and a $2d\rightarrow2d\rightarrow d_p$ MLP to predict standardized proprioception. Both heads use GELU between their linear layers.

\paragraph{GDRF and input embeddings.}
GDRF updates each chunk $h$ using Eq.~\ref{eq:action_feedback}. Its input concatenates the action token, pooled prediction $\overline{\widehat Z}_h^{(r)}\in\mathbb R^d$, goal discrepancy $\overline G-\overline{\widehat Z}_h^{(r)}\in\mathbb R^d$, and encoded proprioception. Without the batch dimension, these form $U_h^{(r)}\in\mathbb R^{4d}$; the full sequence has shape $U^{(r)}\in\mathbb R^{H\times4d}$. The pooled goal $\overline G$ is shared across chunks, and $F_r$ maps these concatenated features to a residual correction.

Within each stage, $F_r$ shares weights across chunks and consists of LayerNorm, a $4d\rightarrow768$ linear layer, GELU, and a $768\rightarrow d$ linear layer. The proprioception encoder $E_p$ is a $d_p\rightarrow d\rightarrow d$ MLP with GELU. A sigmoid gate $\sigma(\gamma_r)$ scales the update. Each stage has one scalar $\gamma_r$, initialized to $-2$ and shared across chunks and features. Time, modality, and validity embeddings mark history positions, input sources, and repeated padding frames.

\subsection{Training objectives and optimization}
\subsubsection{Loss definitions and weights}
Let $A^*$, $Z^*$, and $P^*$ denote demonstrated actions, encoded future scenes, and standardized future proprioception. We choose a goal at chunk $g$ by sampling uniformly from the five chunks, then replacing the choice with the final chunk with probability 0.5. The final chunk is therefore selected with probability 0.6. We write $\nu(A)=(A-\mu_a)/\sigma_a$ for action standardization, $\sg$ for stop-gradient, and $\langle\cdot\rangle_g$ for averaging valid entries up to chunk $g$.

The recorded-action terms are
\begin{align}
\mathcal L_{\rm scene}^{\rm rec}&=\mse(W_Z(M,Z_0,A^*),\sg(Z^*)),\label{eq:recorded_prediction}\\
\mathcal L_{\rm prop}^{\rm rec}&=\slone(W_P(M,Z_0,A^*),\sg(P^*)).
\end{align}
Both terms use all five chunks. The policy terms $\mathcal L_{\rm scene}^{\pi}$ and $\mathcal L_{\rm prop}^{\pi}$ use the same targets and metrics, with predictions conditioned on $A^{(4)}$ and averaged only through $g$.

$W_Z$ and $W_P$ are the world model's scene and proprioception outputs. The recorded-action losses train it on observed action--outcome pairs. The policy-conditioned losses instead encourage the proposed plan to predict the demonstrated future. This future is an alignment target, not an observation collected by executing the proposal. We stop gradients through the goal encoding $G$, but retain them through both feedback paths so that later-stage losses can train earlier proposal updates.

The final and intermediate action losses are
\begin{align}
\mathcal L_{\rm act}&=\langle\slone(\nu(A^{(4)}),\nu(A^*))\rangle_g,\\
\mathcal L_{\rm int}&=\frac13\sum_{r=1}^3\langle\slone(\nu(A^{(r)}),\nu(A^*))\rangle_g.
\end{align}
The goal loss is
\begin{equation}
\mathcal L_{\rm goal}=\mse(\widehat Z^{(4)}_g,\sg(G))+
1-\cos(\overline{\widehat Z}^{(4)}_g,\sg(\overline G)).
\end{equation}
The three core objectives in Eq.~\ref{eq:training_groups} have the following weights:
\begin{align}
\mathcal L_{\rm pred}&=\mathcal L_{\rm scene}^{\rm rec}+0.25\mathcal L_{\rm prop}^{\rm rec},\\
\mathcal L_{\rm action}&=2\mathcal L_{\rm act}+0.25\mathcal L_{\rm int},\\
\mathcal L_{\rm align}&=0.5\mathcal L_{\rm scene}^{\pi}+0.15\mathcal L_{\rm prop}^{\pi}
+\mathcal L_{\rm goal}.
\end{align}
\subsubsection{Optimization and task configurations}
\label{app:optimization}
We train a separate model for each task. DINOv2 remains frozen; all other modules are initialized from scratch and optimized jointly. We use AdamW with a peak learning rate of $5\times10^{-5}$, weight decay of $10^{-3}$, gradient clipping at 1, and BF16 mixed precision. A 1\% warmup is followed by cosine learning-rate decay, without gradient accumulation. PushT is trained for 20 epochs with batch size 512. TwoRoom and Cube use batch size 128 and a ten-epoch schedule. The training seed is 3072. History length is three for PushT/Cube and one for TwoRoom. Action/proprioception dimensions are 2/4, 2/2, and 5/19 for PushT, TwoRoom, and Cube, respectively.

\subsection{Evaluation Protocols}
\label{app:evaluation_details}
Tab.~\ref{tab:execution_protocols} summarizes the prediction horizons, execution lengths, goals, and metrics for the three control settings. The following sections specify the data, evaluation settings, and result sources.
\begin{table}[!htbp]
\centering
\caption{WA-JEPA execution and evaluation protocols. Predicted / executed lengths count environment actions. Four internal stages refine the same pending plan without new observations.}
\label{tab:execution_protocols}
\small\renewcommand{\arraystretch}{1.15}\setlength{\tabcolsep}{3pt}
\begin{tabularx}{\linewidth}{@{}lX@{}}
\toprule
Setting & Protocol \\
\midrule
Short horizon & $25$ / up to $25$ actions; $50$ initial--goal pairs per task. Goal: recorded image at $+25$ steps. Metric: physical goal success. \\
\addlinespace
Long horizon (AP) & $25$ / $5$ actions; $128$ standard + $256$ perturbed starts per task. Goal: retrieved segment, source $+5$ steps. Metric: physical goal success. \\
\addlinespace
Full-task Push-T & $5$ / $1$ actions; $100$ rollouts per run. Goal: task goal. Metric: final coverage. \\
\addlinespace
Full-task LIBERO & $5$ / $1$ actions; $10\times10$ rollouts per run. Goal: end of a training demonstration. Metric: task success on LIBERO-Goal. \\
\bottomrule
\end{tabularx}
\end{table}

\subsubsection{Short-horizon goal reaching}
\label{sec:deployment_details}
We follow App.~\ref{app:cem_budget_followup}. History and goal features are reused across refinement stages; new observations enter only after action execution. Success follows the benchmark's physical criteria: in PushT, the Euclidean error of the concatenated agent and block positions must be below 20 pixels and block orientation error below $\pi/9$ radians; TwoRoom requires agent--goal distance below 16 pixels, and Cube requires position error at most 4\,cm without an orientation requirement. The 4/16 TwoRoom/Cube cases initially within tolerance are retained for both methods; success is checked after execution.

PushT uses an episode-level 90/10 split (16,817/1,868 episodes); TwoRoom and Cube use random-window 90/10 splits with normalization statistics from the full source columns. Baseline scores in Fig.~\ref{fig:mainperformance} come from LeWM's reported results, whereas Tab.~\ref{tab:cem_budget_followup} reports our reevaluation of released checkpoints. Error bars in Fig.~\ref{fig:mainperformance} show success-rate variation across training seeds: baseline bars follow the reported statistics, and WA-JEPA uses $\pm2$ percentage points on PushT, based on results from models trained with three different random seeds.

\subsubsection{Long-horizon anchored planning}
\label{app:ap_protocol}
We follow the long-horizon protocol in Sec.~\ref{sec:longgoal}. Baseline success rates come from the AP release~\cite{liu2026aimshort}. AP (LeWM) uses CEM with 300 candidates, 30 iterations, and 30 elites.

We average CUDA-synchronized decision latencies across both start conditions, including retrieval, rendering, preprocessing, and inference but excluding offline encoding and environment steps. Methods run sequentially on an NVIDIA A40 with two CPU threads.

\subsubsection{Full-task adaptation and evaluation}
\label{app:fulltask_protocol}
Both adaptations use five observation frames, a VQ-BeT head, and four proposal stages to refine one five-action chunk ($H=1$, $b=5$). GDRF compares its predicted outcome with the goal; FSCF passes the unpooled future-scene tokens to the next stage. Temporal ensembling combines overlapping predictions into one executed action per decision. We report means and standard deviations across three independently trained models, each evaluated on 100 episodes. For both tasks, the VQ action autoencoder and codebooks are pretrained on action chunks from training demonstrations and frozen during policy training. The action head's code-prediction and continuous-offset branches are trained jointly with the remaining learnable modules using AdamW, with a learning rate of $5\times10^{-5}$, weight decay of $10^{-3}$, 1\% warmup, and cosine decay.

\paragraph{Push-T.}
The policy uses visual observations and proprioception, pooling each frame into eight 192-dimensional scene tokens. Evaluation allows at most 302 actions and measures final coverage ($71\pm2$\%). Training uses four GPUs, 256 windows per GPU (global batch 1,024), and a 400-epoch schedule. L1 action reconstruction and two focal code-prediction losses have weights 10, 5, and 0.5, with focal $\gamma=2$. The final proposal receives full supervision, plus 0.25 times the mean intermediate-proposal loss. Scene and proprioception prediction weights are 1 and 0.25. Trajectory tails repeat the final valid action for action supervision; unavailable future-scene targets are masked.

\paragraph{LIBERO-Goal.}
We use the Patch Policy release of LIBERO-Goal~\cite{liu2023libero}: ten tasks with 50 demonstrations each, split into 45 training and five validation demonstrations per task. Goals are training-demonstration endpoints. Frozen DINOv2-Small features from two views receive camera embeddings and are jointly pooled into eight scene tokens per timestep. The head predicts five seven-dimensional actions; proprioception input and prediction are omitted. Training runs for 400 epochs with 256 windows per GPU and retains intermediate action supervision and recorded-action future-scene prediction with valid-target masking; the action reconstruction weight is 100. Each model is evaluated on ten episodes per task with a 300-action budget, yielding $97\pm1$\% success under the environment criterion.

\paragraph{Baseline results and re-evaluation.}
Scores and available standard deviations for WebSSL (CLS), Patch Policy, ResNet-18 Patch, and OpenVLA-OFT follow Patch Policy~\cite{zhou2026patchpolicy}.
We re-evaluate public Fast-WAM, RynnVLA-002 and DreamVLA checkpoints on the same 100 seeded LIBERO-Goal cases, with the WA-JEPA environment wrapper, 20 settling steps, identical success criteria, and a 300-action budget.
\noindent Fast-WAM achieves 96\% success, retaining native two-view preprocessing and action normalization, 32 predicted actions, ten executed actions, and ten denoising steps. DreamVLA achieves 86\% with its epoch-28 checkpoint. RynnVLA-002 uses its continuous-action result of 96\%. Both scores are single-checkpoint evaluations; cost measurement follows the common BF16 protocol.

\end{document}